\documentclass[letterpaper, 10 pt, conference]{ieeeconf}  

\IEEEoverridecommandlockouts                              
\usepackage{amsmath} 
\usepackage{url}
\usepackage{cite}
\usepackage{algorithm}
\usepackage{xcolor}
\usepackage{calc}
\usepackage{multicol}
\usepackage{accents}
\usepackage{footnote}
\usepackage{float}
\usepackage[mathscr]{euscript}
\usepackage{bigints}
\usepackage{upgreek}
\usepackage{soul} 
\usepackage[percent]{overpic}
\usepackage{bm}
\usepackage{physics}
\usepackage{mathtools}
\usepackage{amssymb} 
\usepackage{multirow}
\usepackage{amstext}
\usepackage{amsfonts}

\usepackage{graphicx}
\usepackage{fullpage}
\usepackage{caption}
\usepackage{subcaption}

\input{marchand.sty}
\input{andreff.sty}

\renewcommand{\frame}[1]{{\cal R}_{#1}}

\usepackage{tikz}

\makeatletter
\usepackage{pict2e,picture}
\pgfkeys{/csteps/inner ysep/.initial=4pt,
    /csteps/inner xsep/.initial=4pt,
    /csteps/inner color/.initial=black,
    /csteps/outer color/.initial=black,
}
\newsavebox\csteps@CBox
\newlength\csteps@XLength \newlength\csteps@YLength \newlength\csteps@YDepth \newlength\csteps@tmplen
\def\csteps@CircledParam#1#2{\sbox\csteps@CBox{#2}%
    \csteps@XLength=\wd\csteps@CBox\advance\csteps@XLength by\pgfkeysvalueof{/csteps/inner xsep}\relax
    \csteps@tmplen=\pgfkeysvalueof{/csteps/inner ysep}\relax
    \csteps@YDepth=\dp\csteps@CBox\advance\csteps@YDepth by 0.5\csteps@tmplen\relax
    \csteps@YLength=\ht\csteps@CBox\advance\csteps@YLength by\dp\csteps@CBox\advance\csteps@YLength by\pgfkeysvalueof{/csteps/inner ysep}\relax
    \typeout{DBG:#2\space X\space\the\csteps@XLength\space Y:\the\csteps@YLength\space D:\the\csteps@YDepth}%
    \raisebox{-#1\csteps@YDepth}{%
    \ifdim\csteps@XLength>\csteps@YLength
    \makebox[\csteps@XLength]{
        \makebox(0,\csteps@YLength){%
            \color{\pgfkeysvalueof{/csteps/outer color}}\put(0,0){\oval(\csteps@XLength,\csteps@YLength)}%
        }%
    \makebox(0,\csteps@YLength){%
        \put(-.5\wd\csteps@CBox,0){\textcolor{\pgfkeysvalueof{/csteps/inner color}}{#2}}%
    }}%
    \else
    \makebox[\csteps@YLength]{%
        \makebox(0,\csteps@YLength){%
            \color{\pgfkeysvalueof{/csteps/outer color}}\put(0,0){\circle{\csteps@YLength}}%
        }%
    \makebox(0,\csteps@YLength){%
        \put(-.5\wd\csteps@CBox,0){\textcolor{\pgfkeysvalueof{/csteps/inner color}}{#2}}%
     }}%
    \fi
    }%
}
\def\Circled#1{\csteps@CircledParam{1}{#1}}
\def\CircledTop#1{\csteps@CircledParam{0}{#1}}
\makeatother
\title{\LARGE \bf Automatic laser steering for middle ear surgery}

\author{Jae-Hun So$^{1}$, J\'er\^ome Szewczyk$^{1}$ and Brahim Tamadazte$^{1}$.
\thanks{This work has been supported by French ANR $\mu$RoCS Project no ANR-17-CE19-0005-04.}
\thanks{$^{1}$Authors are with Sorbonne Universit\'e, CNRS UMR 7222, INSERM U1150, ISIR, F-75005, Paris, France.
        {\tt\small so@isir.upmc.fr, szewczyk@isir.upmc.fr, brahim.tamadazte@cnrs.fr}
        } 
}        
\begin{document}
\maketitle
\thispagestyle{empty}
\pagestyle{empty}
%
\begin{abstract}
This paper deals with the control of laser spot in the context of minimally invasive surgery of the middle ear, e.g., cholesteatoma removal. More precisely, our work concerns with the exhaustive burring of residual infected cells after a primary mechanical resection of the pathological tissues since the latter cannot guarantee the treatment of all the infected tissues, the remaining infected cells cause regeneration of the diseases in 20\%-25\% of cases, which require a second surgery 12-18 months later. To tackle such a complex surgery, we have developed a robotic platform that consists of the combination of a macro-scale system (7 degrees of freedoms (DoFs) robotic arm) and a micro-scale flexible system (2 DoFs) which operates inside the middle ear cavity. To be able to treat the residual cholesteatoma regions, we proposed a method to automatically generate optimal laser scanning trajectories inside the regions and between them. The trajectories are tacked using an image-based control scheme. The proposed method and materials were validated experimentally using the lab-made robotic platform. The obtained results in terms of accuracy and behavior meet perfectly the laser surgery requirements. 
\end{abstract}
%
%
\section{INTRODUCTION}\label{sec.intro}
%
Robotized Minimally Invasive Surgery (MIS) is gaining more attention due to their accuracy in control of robotic tools~\cite{Vitiello2013}. The adaptation of robotic MIS to smaller workspaces (millimetric scale) is also a trend in surgical robotics. In otorhinolaryngology, contrary to medical disciplines, surgeons still largely use conventional tools to perform surgical tasks that require more precision. This is particularly the case for middle ear surgery, which has received less attention than the placement of electrodes in the inner ear which has been the focus of much research effort over the past decades~\cite{bell2013vitro,caversaccio2019robotic,weber2017instrument}. Among the most recurrent pathologies in the middle ear, cholesteatoma is probably the most frequent which has irreversible consequences on hearing, balance, and facial expression~\cite{eugene_cholesteatoma_1965, bordure2005chirurgie}.  

The cholesteatoma in the middle ear is caused by ventilation problem where dead skin cells cannot be ejected from the tympani forming a cyst. This is mainly due to repeated infection of the tympani. This pile of tumorous tissue expands throughout the middle ear and fills the cavity. Without any treatment, the cholesteatoma tissues start eroding delicate organs and outlining bone, i.e., the mastoid. Consequently, this disease provokes feeling of pressure in the ear, dizziness and at later condition, deafness, brain abscess and meningitis. The only treatment of cholesteatoma is surgery to physically rake out the pathological tissue~\cite{eugene_cholesteatoma_1965}. The surgical procedure is based on performing a mastoidectomy, i.e., a large incision hole (i.e., 20-25mm of diameter) to reach the middle ear cavity, so that the physician has direct access to mechanically remove the pathological tissues with adapted surgical tools~\cite{bordure2005chirurgie}. Until now, the main surgical tool for the ablation is a millimeter-sized hook. This procedure has been considered as a standard method for several years despite numerous limitations: complex procedure, invasiveness, non-exhaustive removal of the infected tissues~\cite{stevens2019canal, blanco2014surgical}. Such imprecise and non-exhaustive surgery leads to a second operation in 20-25\% of cases within 12-18 months and even a third look-up.
 
An efficient solution to remove residual infected cells in the middle ear cavity is the use of a surgical laser~\cite{hamilton2005efficacy}. The idea is to point and burn the remaining cholesteatoma cell clusters to avoid any recrudescence. The composition of the cholesteatoma burns more easily and differently from the middle ear healthy tissue, which makes this method clinically interesting with well-controlled risks. Generally, surgical lasers are used in several ways, including as a scalpel for high-accurate cutting of pathological tissues or ablation by burning all the surface of the infected region, in various fields of medicine~\cite{gonzalez2014robot,franco2015needle,chng2015automation }.

Furthermore, employing visual-based control techniques would make laser manipulation automatic without requiring a human operator. Several studies can be found in the vision-based methods literature related to surgical laser ablation. In micro-phonosurgery~\cite{seon2015decoupling, renevier2016endoscopic, andreff2016laser}, the authors worked on a laser control using a visual servoing controller stereo-vision feedback. The targeted system would be integrating 2 cameras and a 2 DoFs actuating mirror which reflects the laser beam to the surface of vocal cords. For ear surgery, \cite{kahrs2008} reports a laser ablation process to perforate a hole and install a cochlear implant. To achieve such task, the ablation area is first monitored by a color camera to detect the boundary layers of the inner ear. Then, the laser source is automatically guided via visual servoing technique using a two-mirror galvanometric scanner. In retinal surgery, \cite{chng2015automation} presents an image processing technique in order to create a "forbidden zone" where the robotic tool must avoid shooting laser spots for safety reasons. This work introduces a shared control technique in which teleoperation control lets the user move the surgical laser for retinal tissues ablation except in the "forbidden zone".

In this paper, we investigated a minimally invasive approach for residual cholesteatoma treatment, in contrast to the current open surgery. To reach the problem of exhaustive removal of the cholesteatoma tissues, we developed a multiple scale robotic solution~\cite{so2020micro} including a 7 DoFs robotic arm equipped with a 2 DoFs flexible tubular robot. The latter is intended to enter the middle ear cavity through a hole with a diameter of 2-3mm in order to burn automatically the residual cholesteatoma by laser. Achieving manual operation inside the middle ear cavity is highly challenging for surgeons~\cite{fichera2021bringing}. This task should be performed with precision and requires considerable time to remove all residual tissue. It is generally performed by surgeon with high expertise and know-how. 

As a consequence, we developed an automatic method of residual cholesteatoma treatment. It consists of an automatic detection and tracking of the residual cholesteatoma, an optimal path generation to run through all the residual cells, as well as image-guided approach to control the laser spot motion during the infected tissue carbonization. 
 
This manuscript is organized as follows: Section~\ref{sec.path} discusses the detection and visual tracking of both the residual cholesteatoma and the laser spot, and finally the generation of the trajectory which allows to automatically scan the cavity filled with residual infected tissues. Section~\ref{sec.servoing} deals with the formulation of an image-based approach to control the laser spot displacements of along the defined trajectory. Section~\ref{sec.experiment} presents the developed experimental robotic platform as well as the evaluation of the proposed method. 
%
\section{RESIDUAL CHOLESTEATOMA and TRAJECTORY GENERATION} \label{sec.path}
%
This section focuses on the residual cholesteatoma detection and tracking as well as the generation of the path which will allow the laser to be moved to cover all areas of infected tissue. Note that residual cholesteatoma is the debris that remains in the middle ear cavity after the surgeon has resected the large portions of infected tissues. The residual cholesteatoma tissues are relatively small, randomly distributed in the middle ear, and have whiter appearance than the rest of the healthy tissue. The surgeon then proceeds to the meticulous carbonization of the residual cells aggregates using a surgical laser, usually of the KTP (Potassium titanyl phosphate) type with a diameter of the laser varying from 50$\mu$m to 200$\mu$m~\cite{fichera2016learning}. 
\subsection{Cholesteatoma and Laser Spot Tracking}
%
The first step of the proposed method concerns with the detection of the residual cholesteatoma regions and the laser spot and their visual tracking over time. To do this, we evaluated different open-source visual tracking algorithms and the one that seems to be the most efficient and the most adapted for this task is the Channel and Spatial Reliability Tracker (CSRT)~\cite{lukezic2017discriminative} available in OpenCV library. It is based on the estimation of the reliability using the properties of a constrained least-squares solution in the filter design surrounding the region of interest suitable for tracking. Therefore, the channel reliability scores are used for weighting the per-channel filter responses in the object localization process. Note that the filter support can be adjusted during the tracking which allows to enlarge the search region and improves tracking of non-rectangular objects. This means that this approach is perfectly suitable for our residual cholesteatoma tracking problem, which varies in shape and size from one sample to another.
\begin{figure}[!h]
    \centering
    \includegraphics[width =\columnwidth]{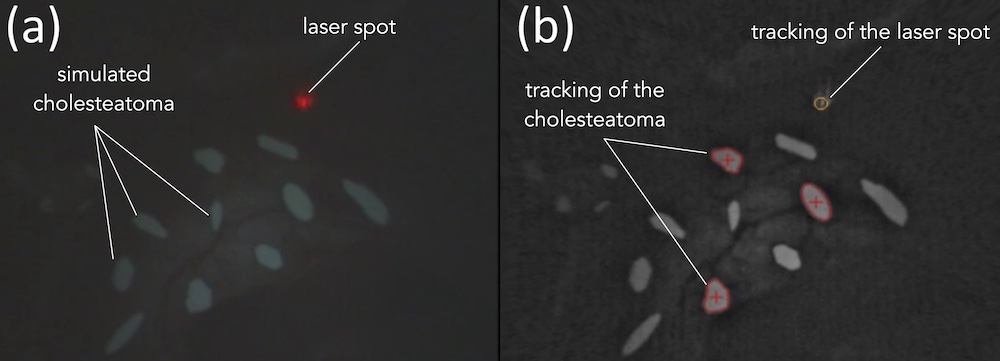}
    \caption{Example of visual tracking of residual cholesteatoma and laser spot.}
    \label{fig.simu.choles}
\end{figure}
 
The tracking algorithm then provides the bounding rectangle of each region of cholesteatoma as well as the laser spot. Now, the challenge is to find the shortest path that connects the different residual cholesteatoma regions to each other.

\subsection{Intra-regions Trajectory Generation}
%
The whole trajectory of the laser spot includes linear portions between successive regions, plus a portion corresponding to the intra-region scanning strategy.

\begin{figure}[!h]
    \centering
    \includegraphics[width = .75\columnwidth]{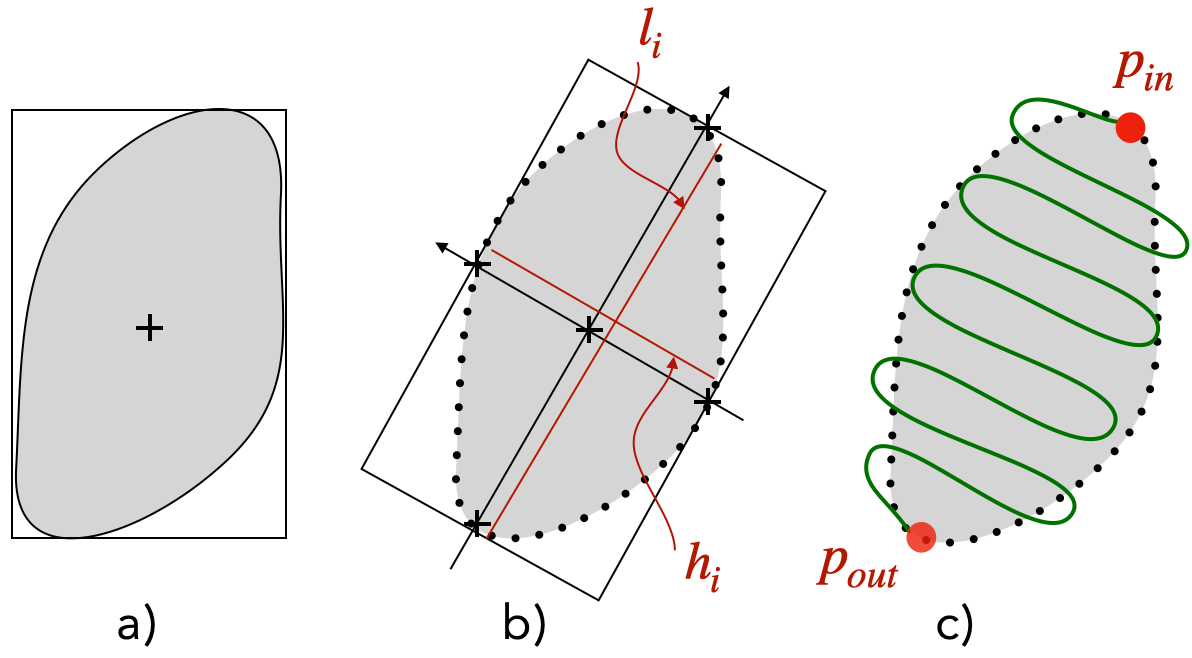}
    \caption{Intra-regions path generation step. (a) Extraction of contour points list (b) Extraction of geometrical parameters (c) Drawing of sinusoidal-like path to cover the "residual" zone.}
    \label{fig:acp}
\end{figure}

The intra-region scanning path is intended to move the laser spot cover the entire area of a residual cholesteatoma. The coverage will of course depend on the geometrical property of this area. As presented earlier, the visual tracking method of residual cholesteatoma regions provides the position, the bounding box as well as the contour-points list of each cholesteatoma region. Based on this, Principal Components Analysis (PCA) is used to determine the input points $\mathbf{p}_{in} = (\mathbf{p}_{in}, \mathbf{y}_{in})^\intercal$ and the output points $\mathbf{p}_{out} = (\mathbf{x}_{out}, \mathbf{y}_{out})^\intercal$. These points correspond to the intersection between the first principal axis found by the PCA method and the contour of a given region (see Figure~\ref{fig:acp}).

Let us consider the contours points that delimit one cholesteatoma region defined by ${p_i} = [{p_1},{p_2},...,{p_N}]$ with $N$ is the total number of contours-points of a given region. The reprojection $p'_i$ of the contour-points ${p_i}$ according to the first two principal components is given by: 
\begin{equation}
   p'_i = \mathbf{A}(p_i - p_c)
    \label{eqn:pca}
\end{equation}
where $p_{c} = (x_{c}, y_{c})^\intercal$ is the centroid of the considered region, computed as:
\begin{equation}
       p_{c}=\frac{1}{N} \sum_{i=1}^{N} {p}_{i},
\end{equation}
and $\mathbf{A} \in \mathbb{R}^{2 \times 2}$ is the transformation matrix from the original representation to the PCA one. It is obtained by the computation of the eigenvectors of the covariance matrix $\mathbf{C}$ given by:
\begin{equation}
	\mathbf{C} = 
	\left(\begin{array}{cc}
    {var}(x_i) & {cov}(x_i, y_i) \\
    {cov}(y_i, x_i) & {var}(y_i)
    \end{array}\right)
\end{equation}
where $var$ and $cov$ represent the variance and the covariance, respectively.

\begin{figure}[!h]
    \centering
    \includegraphics[width = 0.8\columnwidth]{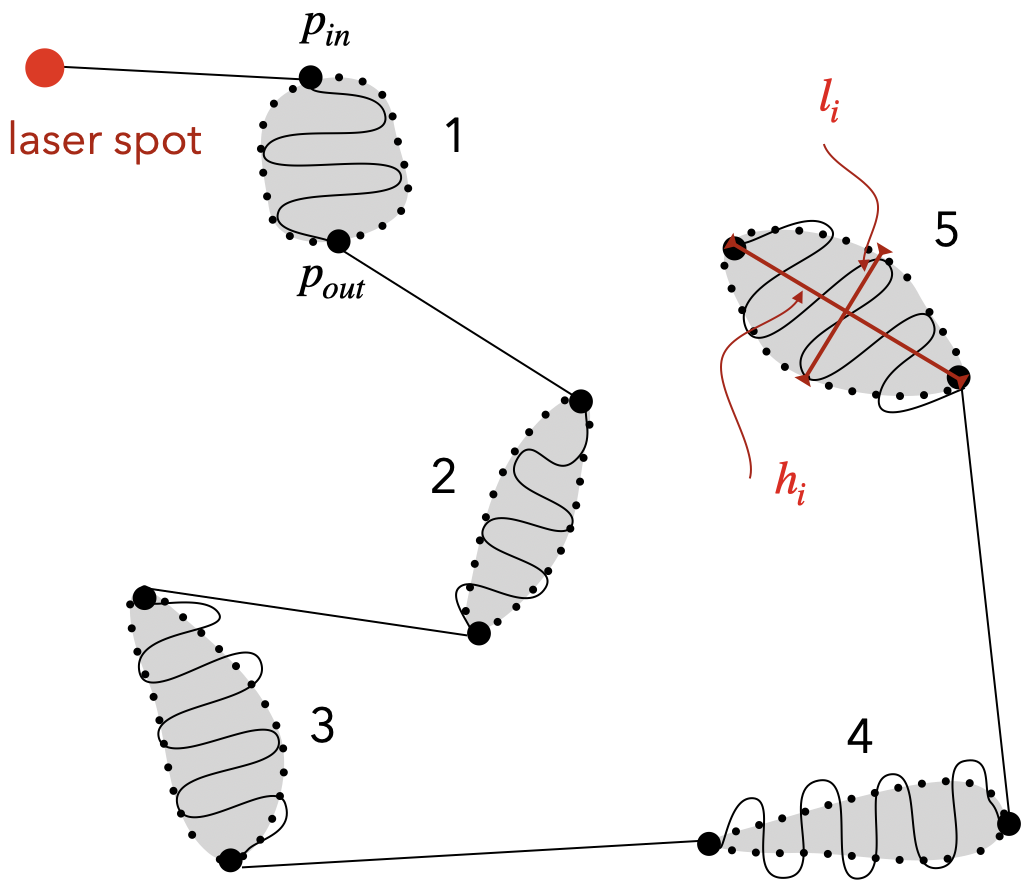}
    \caption{Illustration of the generation of an scanning trajectory for the treatment of 5 residual cholesteatoma regions.}
    \label{fig.traj}
\end{figure}

Having the input and the output points $\mathbf{p}_{in}$ and $\mathbf{p}_{out}$, the length $l_i$ and the width $h_i$ of each cholesteatoma region, it is possible to define the scanning curve that allows covering the whole surface of the residual cholesteatoma region. Different scanning curves can be candidates. In this work, we opted for a simple sinusoidal curve defined by the following relationship: 
\begin{equation}
   y' = h_i\sin{\Bigg(\frac{2\pi}{d_{laser}}} x'\Bigg) ~~~~ \text{ with }  x' \in [0,l_i]
\end{equation}
where $d_{laser}$ is the diameter of the laser spot (Fig.~\ref{fig.traj}).  

Finally, the coordinates of each sinusoidal curve $\tilde{p}_i = (\tilde{x}',\tilde{y}')$ are expressed in the image frame $\frame{I}$ by the inverse of (\ref{eqn:pca}): 
\begin{equation}
    \tilde{p}_i = \mathbf{A}^\intercal \tilde{p}'_i+p_c
\end{equation}
%
\subsection{Inter-regions Trajectory Generation}
\begin{figure}[!h]
    \centering
    \includegraphics[width = \columnwidth]{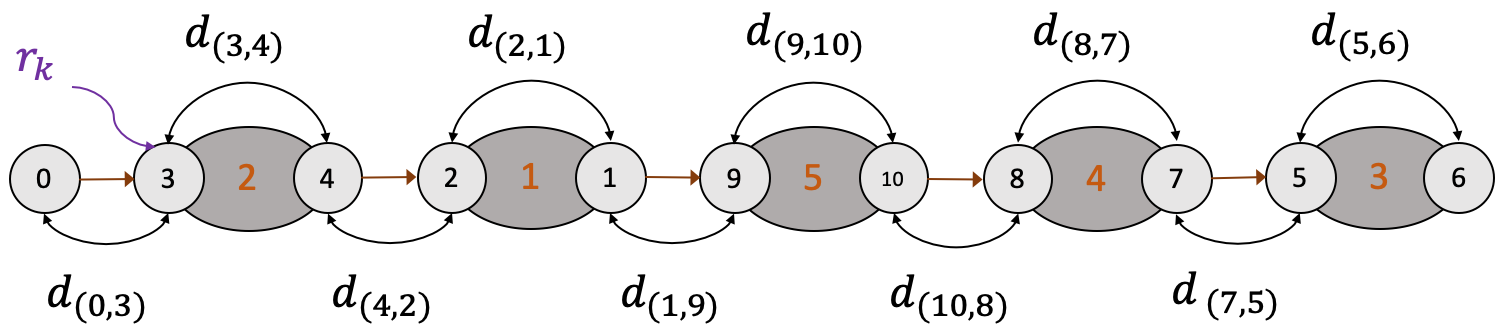}
    \caption{Construction of the TSP problem as a graph of nodes. Here, considered nodes are the laser spot position and the pair of points that forms the input and output of each cholesteatoma region (5 in total).}
    \label{fig.tsp}
\end{figure}

%
The last step of the design of the whole scanning path consists of the connection between the cholesteatoma regions and the path should be as short and accurate as possible. This can be associated with a typical optimization problem, widely studied in transportation theory, known as Traveling Salesman Problem (TSP)~\cite{lawler1985traveling}. In this study, we proposed an adaptation of the traditional TSP algorithm by adding a new condition during the computation of the scanning path of a set of locations. The considered set of 2D positions includes: the current position of the laser spot $p_l = (x_l, y_l)$ and a pair of points that forms the input and output points of each region, i.e., $p_{in} = ({x}_{in}, {y}_{in})$ and ${p}_{out} = ({x}_{out}, {y}_{out})$, respectively. At this stage, the TSP algorithm does not yet know the input and output points of each region. To formalize the optimization problem (Figure~\ref{fig.tsp}), we consider ($2n+1$) "nodes" including the current laser position and the input and output positions of all the $n$ regions.
To solve the optimization problem, we start by constructing the whole graph of nodes. This is formalized by the following method:
\begin{itemize}
    \item Each region is assigned a number $k$, $k \in [1, 2, \dotsm, n]$;
    \item For each region, the two extremities (i.e., the future input and output points are assigned the number $r_k$ via equation \eqref{eq:rknum};
    \item Systematically, a "0" represents the initial location of the laser spot.
\end{itemize}

For example, for the region $\Circled{~2~}$, the associated nodes are $\Circled{3}$ and $\Circled{4}$ (see Fig.~\ref{fig.tsp}).
\begin{equation}
    \begin{cases}
    r_{k} = 2k-1 \\
    r_{k+1} = 2k  
    \end{cases} 
    \text{for} \ \ k = 1, \dotsm, n
    \label{eq:rknum}
\end{equation}
 
The next step is to construct a sequence of nodes $S = (s_0,s_1,s_2,\dotsm,s_{2n})$ defining the laser scanning path. For that, for each pair of nodes, a Cartesian distance $d$ is computed. The algorithm then attempts to find the sequence that offers the shortest path to connect all nodes. This minimum distance $d_{\min}$ is obtained as follows: 
\begin{equation}
    d_{\min} = \min{\sum_{i=0}^{2n} d_{(s_i, s_{i+1})}}
\label{eqn.mintsp}
\end{equation}

Based on the traditional TSP problem, we added supplementary conditions corresponding to the study case, i.e., 1) the first node $s_0$ is always the laser location, 2) successive pairs of nodes $\big((s_1, s_2), (s_3, s_4), ...\big)$ are belong to the same region $k$, and 3) the absolute difference value between the assigned number of each pair of nodes must be 1. These conditions can be formalized as follows:
\begin{eqnarray}
\nonumber
\left\{\begin{matrix}
 s_0 & =  & 0 &  ~\\ 
 \frac{1+s_{2i-1}+s_{2i}}{4} & = & k &  ~~~~~ \text{for} ~~i = 1, 2, \cdots, n \\ 
 \lvert s_{2i-1}-s_{2i} \rvert & = & 1& ~~~~~ \text{for} ~~i = 1, 2, \cdots, n  
\end{matrix}\right.
\end{eqnarray}

An example of the result of this algorithm can be seen in Fig.~\ref{fig.tsp}. In this example, five regions are considered making a total of 11 nodes organized as a sequence
$S = (0,3,4,2,1,9,10,8,7,5,6)$, which gives the shortest trajectory to cover the simulated residual cholesteatoma.

At this stage of development, the entire laser path in the middle ear cavity is constructed. It includes sinusoidal parts where the surgical laser is activated to treat the cholesteatoma tissues and other straight-line segments parts where only the aiming (visible) laser  is activated. 
\section{LASER STEERING ALONG A TRAJECTORY} \label{sec.servoing}
%
Several control strategies can be considered to control the laser motion. Nevertheless, the requirement of laser surgery (e.g., accuracy), a closed-loop solution is preferred such as image-based visual servoing (IBVS) strategy. Here, the generated trajectories are used as the desired positions of the laser spot during the cholesteatoma treatment. 
%
%
According to~\cite{chaumette2006visual}, the aim of a visual servoing is to control the motion of a dynamic system in order to allow a set of geometric visual features $\ms$ ($\ms \in \mathbb{R}^k$) defining a robot pose $\mathbf{r}(t) \in SE(3)$ \Big(i.e., $\ms = \ms\big(\mathbf{r}(t)$\big)\Big) to reach a set of desired ones $\msd$ ($\msd \in \mathbb{R}^k$), which represent the desired positions of the laser spot along a predefined trajectory, by minimizing a visual error given by:
\begin{equation}
\me = \ms \big(\mathbf{r}(t)\big) - \msd
\label{eq.e}
\end{equation}

The time-variation of $\mathbf{s}$ is linked to the velocity twist $\tc = (v_x \quad v_y \quad v_z \quad \omega_x \quad \omega_y \quad \omega_z)^\top$ of the camera frame by $\dot{\ms} = \mLs \tc$, where $\mLs \in \mathbb{R}^{k \times 6}$ is commonly called the \emph{interaction matrix} (or \emph{Jacobian image} in certain papers). The expression of $\mLs$ is given in~\cite{chaumette2006visual}. 

From (\ref{eq.e}), the variation of the visual error $\me$ due to the visual sensor velocity is $\dot{\me} = \mLs \tc - \dot{\ms}^*$ (here, $\dot{\ms}^* = 0$). If one wants to ensure an exponential decoupled decrease of the error $\me$ ($\dot{\me} = -\lambda \me$),  it becomes possible to express the velocity tensor as follows:
\begin{equation}
\tc = -\lambda \widehat{\mLs}{^+} \big(\ms(t) - \msd \big)
\label{equ.controller}
\end{equation}
where $\lambda$ is a positive gain and $\widehat{\mLs}^{+}$ $\Big( \widehat{\mLs}^{+} = \big(\widehat{\mLs}^\top \widehat{\mLs}\big)^{-1}\widehat{\mLs}^\top \Big)$ is the \emph{Moore-Penrose} pseudo-inverse of the approximated interaction matrix denoted $\widehat{\mLs}$.

Note that the laser is mounted on a flexible and actuated tool which is in turn mounted on the Panda robot arm as shown in Fig.~\ref{fig:setup}(a). The scene is viewed by an external endoscopic system (Fig.~\ref{fig:setup}(b)). The constructed system is an eye-to-hand configuration, the relation between the robot velocity $\mathbf{\dot q}$ and the camera one $\tc$ is expressed by:
\begin{equation}
\mathbf{\dot q} = - ^{e}{\mathbf{K}}_{b} ~ ^b\mathbf{V}_c ~ ^c\tc_c 
\label{eq.EyetoHandIntera}
\end{equation}
where $^e\mathbf{K}_b$ is the inverse kinematic Jacobian matrix of the robot in the base frame $\frame\cal{~_b}$, and $^b\mathbf{V}_c$ is the transformation matrix associated with the velocity which is shifted from $\frame \cal{_b}$ to $\frame \cal{_c}$ (i.e., the camera frame). 
%
\section{EXPERIMENTAL VALIDATION} \label{sec.experiment}
%
\subsection{Experimental Setup}
%
\begin{figure}[!h]
   \centering
    \includegraphics[width =  .9\columnwidth]{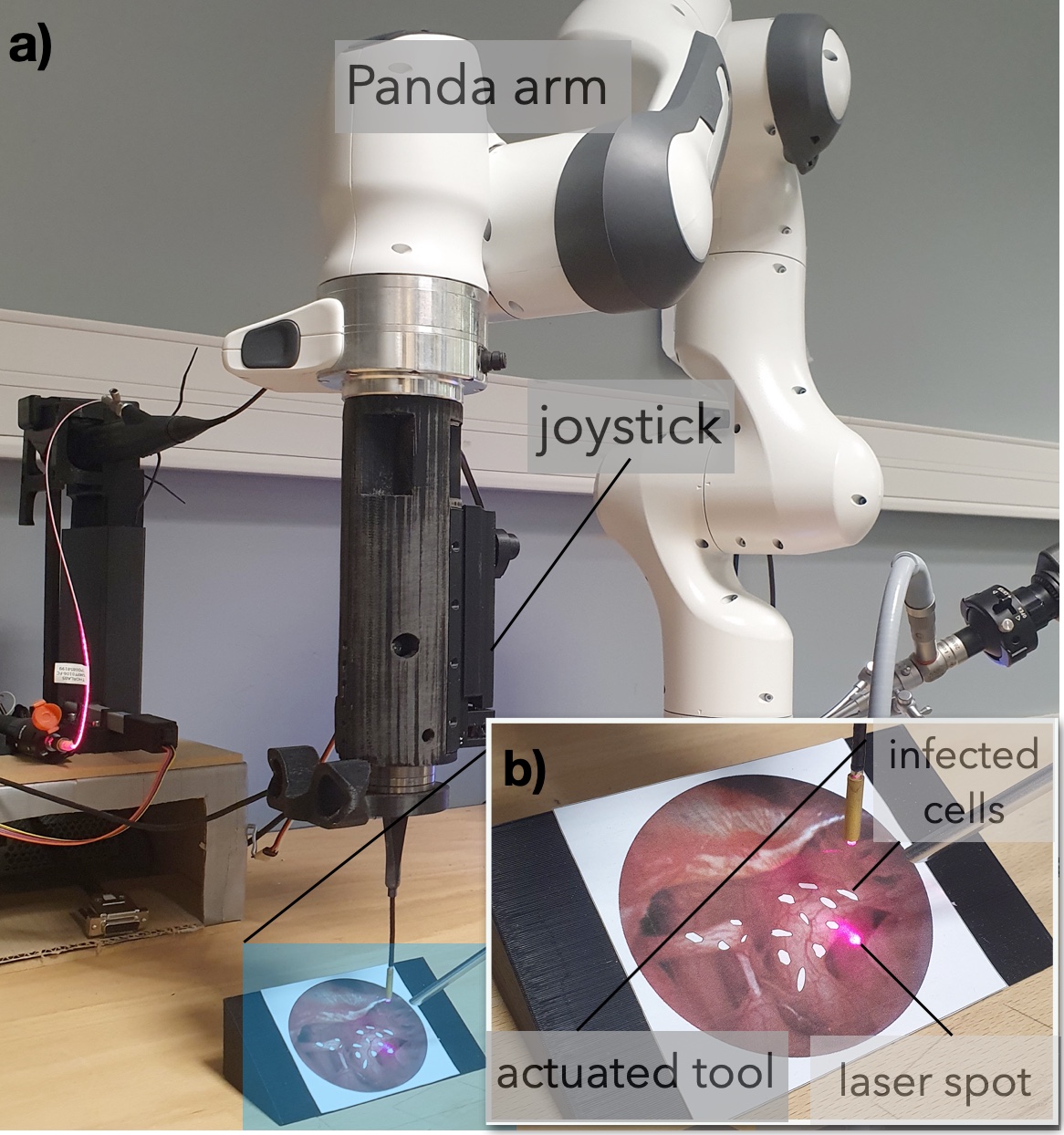}
    \caption{Photography of the designed robotic setup: a) is a global view of the whole robotic system and b) a zoom-in on the distal part of the flexible tool embedding the laser source and the scene used to mimic the middle ear cavity and the residual cholesteatoma.}
    \label{fig:setup}
\end{figure}

The proposed methods have been experimentally validated using the developed robotic station depicted in Fig.~\ref{fig:setup}(a). The latter includes a Panda 7~DoFs robotic arm, in which is attached a 2~DoFs actuated and flexible mechanism. The fiber laser source is fixed along the flexible system. An external rigid endoscope (from KARL STORZ) equipped with an integrated light source is used to visualize the scene. The endoscopic camera offers 768$\times$576 pixels images at the frame-rate of 25 images/second. 

The developed image processing, path generation, and control methods were implemented in Robot Operating System (ROS) framework. Then, we set up an evaluation scenario that mimics the residual cholesteatoma's laser ablation after a mechanical resection of the significant parts of infected tissues. To do so, a simulating image depicting the middle ear cavity and the remaining cholesteatoma tissue after the first phase of intervention is added to the setup at a location consistent with the middle ear cavity anatomy. Note that the experimental setup is scaled up with a ratio of 1:3 (Fig.~\ref{fig:setup}(b)).
%
\subsection{Results Analysis and Discussion}
Various trials with different numbers and positions of simulated residual cholesteatoma were conducted. Three main points are evaluated in this phase: the performance of the vision-based control of the laser spot during the path following, the generation of the optimized path, and the coverage of residual cholesteatoma by the laser spot. 
\begin{figure}[!h]
    \centering
    \includegraphics[width=\columnwidth]{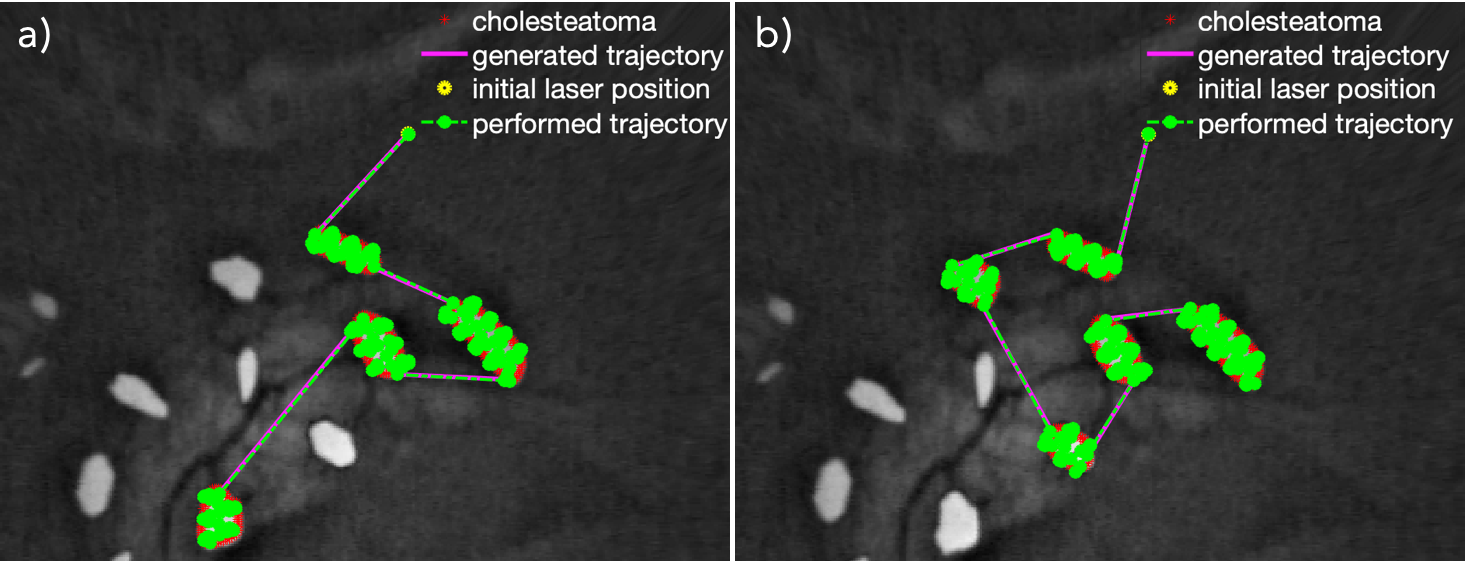}
    \caption{Some examples of performed trajectories (in green): a) with 4 cholesteatoma regions and b) with 5 residual cholesteatoma.}
    \label{fig.traj-done}
\end{figure}
	    
The performance of the vision-based steering method is analyzed during the path following. The behavior of the laser spot motion as well as the accuracy of the image-based path tracking is evaluated experimentally. Figure~\ref{fig.traj-done} shows example of generated (in magenta) and performed (in green) paths. As can be highlighted, both performed inter-region (linear portions) and intra-region (sinusoidal portions) paths are almost superposed with the desired ones.

\subsubsection{Accuracy}
The good path following accuracy is confirmed by the tracking errors for different achieved paths summarized in Table~\ref{tab.error}. The average error $\bar{e}$ for each evaluation is estimated to 85$\mu$m (with a STD of about 25$\mu$m) regardless of the number of considered residual cholesteatoma regions, \textit{e.g.}, 3, 4 or 5. The mean error value is in accordance with the threshold set as 1.0 px (120 $\mu$m), wherein the desired position is considered as reached when the error is below the predefined threshold.
\begin{table}[!h]
\small
\centering
\resizebox{0.9\columnwidth}{!}{%
\begin{tabular}{|l|c|c|c|c|}
\hline
Test & \begin{tabular}[c]{@{}c@{}} Number of\\ Regions\end{tabular} & \begin{tabular}[c]{@{}c@{}} Mean Error ($\bar{e}$)\\ ($\mu$m)\end{tabular}  & STD ($\mu$m) & Cover (\%) \\ \hline \hline
1    & 3                                                          & 83.72                    & 27.09  &  99.38 \\ \hline
2    & 3                                                          & 83.33                    & 26.98  &  97.77 \\ \hline
3    & 4                                                          & 77.90                    & 27.87  &  98.71 \\ \hline
4    & 4                                                          & 81.40                    & 27.63  &  96.62 \\ \hline
5    & 5                                                          & 86.05                    & 26.51  &  98.01 \\ \hline
6    & 5                                                          & 87.63                    & 26.59  &  99.00 \\ \hline
7    & 3                                                          & 89.78                    & 22.21  &  97.82 \\ \hline
8    & 3                                                          & 90.17                    & 21.35  &  97.77 \\ \hline
9    & 4                                                          & 87.31                    & 25.60  &  98.37 \\ \hline
10   & 4                                                          & 90.60                    & 22.79  &  97.82 \\\hline \hline
 ~   & average                                               &   85.78 $\mu$m  &   25.46 $\mu$m &  98.12\% \\\hline
\end{tabular}}
\caption{Accuracy analysis for the paths generated for different scenarios with 3, 4 or 5 residual cholesteatoma regions.}
\label{tab.error}
\end{table}

\begin{figure}[!h]
    \centering
    \includegraphics[width=.75\columnwidth]{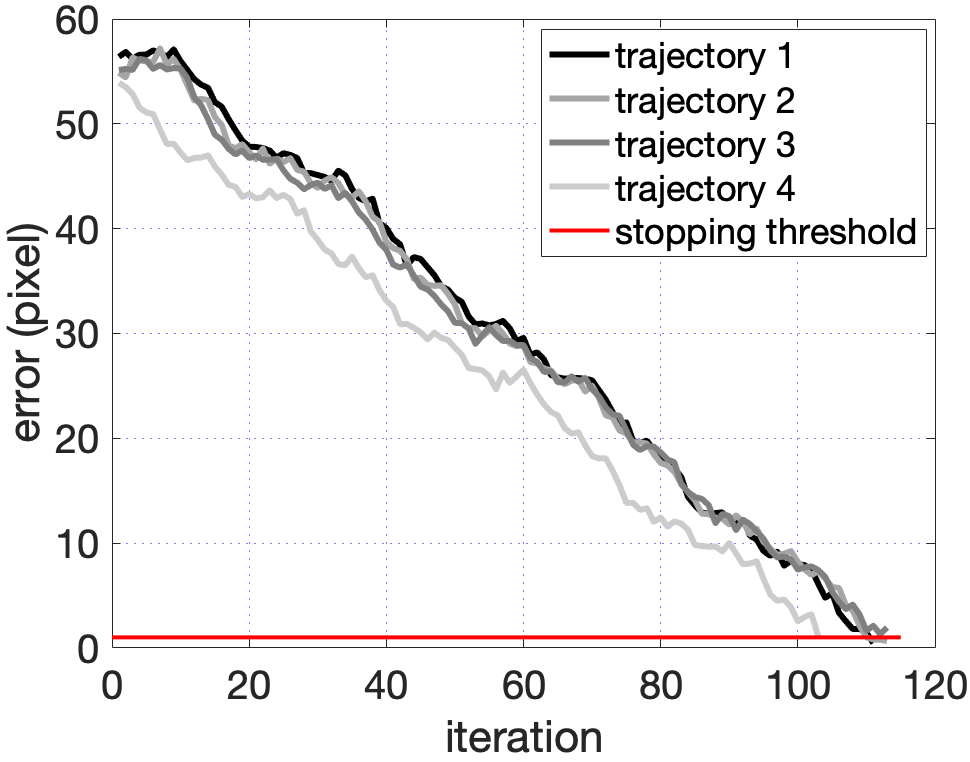}
    \caption{Example of error decay versus iterations (each iteration takes 100ms) for different achieved trajectories. The convergence criteria is fixed to 1 pixel corresponding to 0.12mm.}
    \label{fig.norm_error}
\end{figure}

The norm of the Cartesian error decay was recorded for several performed trajectories and plotted in Fig.~\ref{fig.norm_error}. The regulation of the error was linear instead of the expected traditional exponential decay. It can be explained by the motion of the actuated fibroscope that carries the laser source. Note the fibroscope is actuated by wires whose motors are located at 100cm from the distal part. Hence, small movements of the distal part of the fibroscope (corresponding the sinusoidal-like portions) are limited by the friction of the actuating wires.

\subsubsection{Optimal path generation}
\begin{figure}[!h]
    \centering
    \includegraphics[width=\columnwidth]{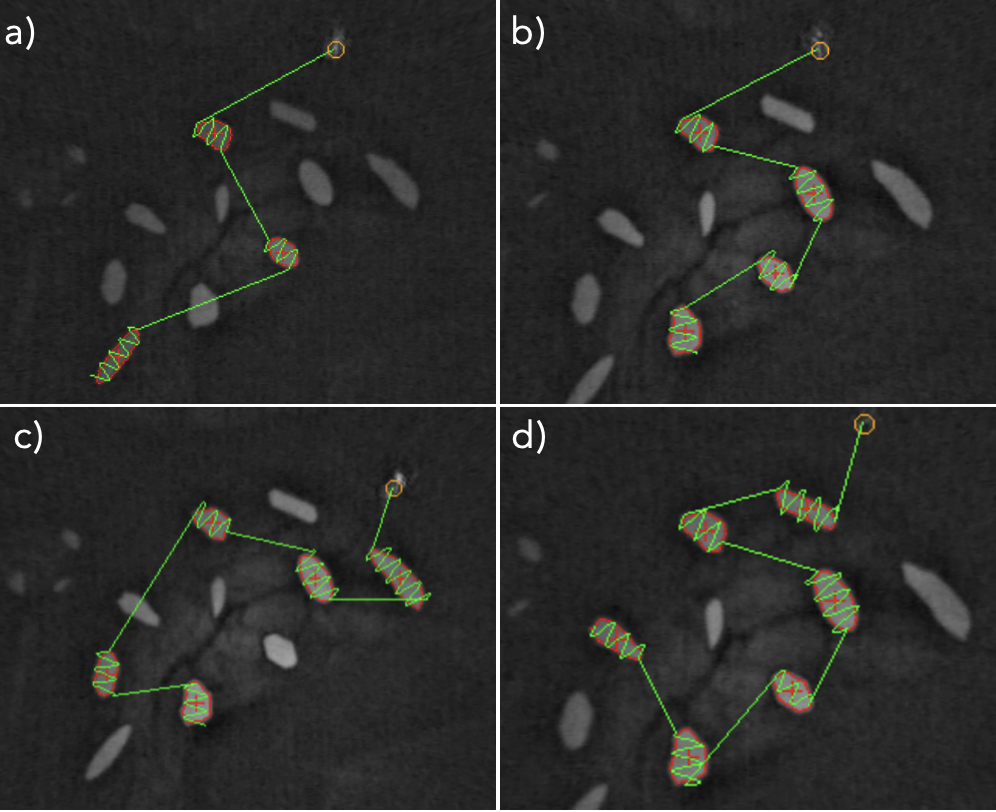}
    \caption{Examples of generated scanning trajectories: a) 3 cholesteatoma regions to burn, b) 4 regions, c) 5 regions, and d) 6 regions.}
  \label{fig.residuels}
\end{figure}

The detection and visual tracking of the remaining infected tissues as well as the path generation, were also evaluated. Different numbers and positions of residual cholesteatoma were considered. The generated scanning paths (inter and intra-region) are the results of the developed augmented TSP algorithm in which the minimal distance cost is retained from every possible path. Figure~\ref{fig.residuels} shows only the retained minimized paths which fit perfectly the optimal scanning problem in terms of path length minimization. 
	    
\subsubsection{Laser Coverage}
The last evaluation criterion consisted of the ability to sufficiently cover the cholesteatoma regions with the laser spot to carbonize the infected tissue completely. This parameter is calculated as an overlap rate of all positions of the laser spot over the total surface of the treated region. The laser spot diameter is fixed to $d_{laser}=$ 5 pixels, equivalent to 600 $\mu$m ). For the ten tests reported in Table~\ref{tab.error}, the average covering rate is estimated to be 98.12\%, which almost perfectly meets the specifications of laser surgery. This rate could also be improved by changing the curve parameters, i.e. by changing the periodic parameter of the sine wave or studying other curves (path) that could more efficiently cover the cholesteatoma regions.
%
\section{CONCLUSION} \label{sec.conclu}
%
In this paper, we discussed the development of a robotic solution for minimally invasive surgery of the middle ear. It aimed at the treatment of the remaining residual cholesteatoma inside in the middle ear cavity using a surgical laser. This type of treatment is considered as a gold standard in residual cholesteatoma treatment. To do this, we developed an image-based controller and a scanning trajectory generation method that allowed to control the displacement of the laser spot within the middle ear as well as the scanning of the whole surface of each cholesteatoma region.  

The proposed methods and materials were successfully evaluated experimentally using a robotic setup. The obtained results are convincing in term of positioning accuracy and laser motion.  

Our future work will involve the evaluation of the proposed methods in experimental conditions similar to those of a middle ear surgery. Additionally, we are currently working on an original micro-robotic system that will integrate miniature surgical tools, a surgical laser and a fiber-optic camera which will replace the current flexible fibroscope.  

\renewcommand{\baselinestretch}{0.99}
%

\begin{thebibliography}{10}
\providecommand{\url}[1]{#1}
\csname url@rmstyle\endcsname
\providecommand{\newblock}{\relax}
\providecommand{\bibinfo}[2]{#2}
\providecommand\BIBentrySTDinterwordspacing{\spaceskip=0pt\relax}
\providecommand\BIBentryALTinterwordstretchfactor{4}
\providecommand\BIBentryALTinterwordspacing{\spaceskip=\fontdimen2\font plus
\BIBentryALTinterwordstretchfactor\fontdimen3\font minus
  \fontdimen4\font\relax}
\providecommand\BIBforeignlanguage[2]{{%
\expandafter\ifx\csname l@#1\endcsname\relax
\typeout{** WARNING: IEEEtran.bst: No hyphenation pattern has been}%
\typeout{** loaded for the language `#1'. Using the pattern for}%
\typeout{** the default language instead.}%
\else
\language=\csname l@#1\endcsname
\fi
#2}}

\bibitem{Vitiello2013}
V.~Vitiello, {S.-L. Lee}, T.~P. Cundy,  \emph{et~al.}, ``{Emerging
  Robotic Platforms for Minimally Invasive Surgery},'' \emph{IEEE R. in
  Biomed. Eng.}, vol.~6, pp. 111--126, 2013.

\bibitem{bell2013vitro}
B.~Bell, N.~Gerber, T.~Williamson, \emph{et~al.}, ``In vitro accuracy evaluation of image-guided robot system for
  direct cochlear access,'' \emph{Otology \& Neurotology}, vol.~34,  pp.
  1284--1290, 2013.

\bibitem{caversaccio2019robotic}
M.~Caversaccio, W.~Wimmer, J.~Anso, \emph{et~al.}, ``Robotic
  middle ear access for cochlear implantation: first in man,'' \emph{PloS One},
  vol.~14, no.~8, p. e0220543, 2019.

\bibitem{weber2017instrument}
S.~Weber, K.~Gavaghan, W.~Wimmer, \emph{et~al.}, \emph{et~al.}, ``Instrument flight to
  the inner ear,'' \emph{Science Robotics}, vol.~2, no.~4, 2017.

\bibitem{eugene_cholesteatoma_1965}
E.~L. Derlacki and J.~D. Clemis, ``Lx congenital cholesteatoma of the middle
  ear and mastoid,'' \emph{Ann. of Otology, Rhinology \& Laryngology},
  vol.~74,  pp. 706--727, 1965.

\bibitem{bordure2005chirurgie}
P.~Bordure, A.~Robier, and O.~Malard, \emph{Chirurgie otologique et
  otoneurologique}.\hskip 1em plus 0.5em minus 0.4em\relax Elsevier Masson,
  2005.

\bibitem{stevens2019canal}
S.~M. Stevens, Z.~A. Walters, K.~Babo, \emph{et~al.}, ``Canal reconstruction mastoidectomy: Outcomes comparison following
  primary versus secondary surgery,'' \emph{The Laryngoscope}, vol. 129,
  no.~11, pp. 2580--2587, 2019.

\bibitem{blanco2014surgical}
P.~Blanco, F.~Gonz{\'a}lez, J.~Holgu{\'\i}n, \emph{et~al.}, ``Surgical
  management of middle ear cholesteatoma and reconstruction at the same time,''
  \emph{Colombia M{\'e}dica}, vol.~45, no.~3, pp. 127--131, 2014.

\bibitem{hamilton2005efficacy}
J.~W. Hamilton, ``Efficacy of the ktp laser in the treatment of middle ear
  cholesteatoma,'' \emph{Otology \& Neurotology}, vol.~26, no.~2, pp. 135--139,
  2005.

\bibitem{gonzalez2014robot}
J.~Gonzalez-Martinez, S.~Vadera, J.~Mullin, \emph{et~al.},``Robot-assisted stereotactic laser
  ablation in medically intractable epilepsy: operative technique,''
  \emph{Operative Neurosurgery}, vol.~10, no.~2, pp. 167--173, 2014.

\bibitem{franco2015needle}
E.~Franco, D.~Brujic, M.~Rea, \emph{et~al.}, ``Needle-guiding
  robot for laser ablation of liver tumors under mri guidance,''
  \emph{IEEE/ASME Trans. on Mech.}, vol.~21, no.~2, pp. 931--944,
  2015.

\bibitem{chng2015automation}
C.-B. Chng, Y.~Ho, and C.-K. Chui, ``Automation of retinal surgery: A shared
  control robotic system for laser ablation,'' in \emph{IEEE Int.
  Conf. on Inf. and Auto.}, 2015, pp. 1957--1962.

\bibitem{so2020micro}
J.-H. So, B.~Tamadazte, and J.~Szewczyk, ``Micro/macro-scale robotic approach
  for middle ear surgery,'' \emph{IEEE Trans. on Med. Rob. and
  Bio.}, vol.~2, no.~4, pp. 533--536, 2020.

\bibitem{fichera2021bringing}
L.~Fichera, ``Bringing the light inside the body to perform better surgery,''
  \emph{Science Robotics}, vol.~6, no.~50, p. eabf1523, 2021.

\bibitem{seon2015decoupling}
J.-A. Seon, B.~Tamadazte, and N.~Andreff, ``Decoupling path following and
  velocity profile in vision-guided laser steering,'' \emph{IEEE Trans.
  on Rob.}, vol.~31, no.~2, pp. 280--289, 2015.

\bibitem{renevier2016endoscopic}
R.~Renevier, B.~Tamadazte, K.~Rabenorosoa, \emph{et al.},
  ``Endoscopic laser surgery: Design, modeling, and control,'' \emph{IEEE/ASME
  Trans. on Mech.}, vol.~22, no.~1, pp. 99--106, 2016.

\bibitem{andreff2016laser}
N.~Andreff and B.~Tamadazte, ``Laser steering using virtual trifocal visual
  servoing,'' \emph{The Int. J. of Rob. Res.}, vol.~35,
  no.~6, pp. 672--694, 2016.
  
\bibitem{kahrs2008}
L. A. Kahrs,  \emph{et al.}, "Visual servoing of a laser ablation based cochleostomy." \emph{Medical Imaging}, , p. 69182C, 2008.

\bibitem{lukezic2017discriminative}
A.~Lukezic, T.~Vojir, L.~Cehovin~Zajc, J.~Matas, and M.~Kristan,
  ``Discriminative correlation filter with channel and spatial reliability,''
  in \emph{IEEE Conf. on Comp. Vis. and Pat. Recog.}, 2017, pp. 6309--6318.

\bibitem{fichera2016learning}
L.~Fichera, ``Learning the temperature dynamics during thermal laser
  ablation,'' in \emph{Cognitive Supervision for Robot-Assisted Minimally
  Invasive Laser Surgery}, 2016, pp. 43--62.

\bibitem{lawler1985traveling}
E.~L. Lawler, ``The traveling salesman problem: a guided tour of combinatorial
  optimization,'' \emph{Wiley-Interscience Series in Discrete Mathematics},
  1985.

\bibitem{chaumette2006visual}
F.~Chaumette and S.~Hutchinson, ``Visual servo control. i. basic approaches,''
  \emph{IEEE Rob. \& Auto. Mag.}, vol.~13, no.~4, pp. 82--90,
  2006.
\end{thebibliography}

\end{document}